\documentclass[conference]{IEEEtran}
\usepackage{cite}
\usepackage{amsmath,amssymb,amsfonts}
\usepackage{algorithmic}
\usepackage{graphicx}
\usepackage{textcomp}
\usepackage{xcolor}
\usepackage{placeins}
\usepackage{float}
\usepackage{fancyhdr}
\usepackage[hidelinks]{hyperref}

\begin{document}

\fancypagestyle{firstpage}{%
  \fancyhf{}
  \renewcommand{\headrulewidth}{0pt}
  \renewcommand{\footrulewidth}{0pt}
  \fancyhead[R]{\footnotesize 11th IEEE Asia-Pacific Conference on Computer Science and Data Engineering (IEEE CSDE 2026)\\
  1-3 November 2026, Cox's Bazar, Bangladesh}
  \fancyfoot[L]{\footnotesize 979-8-3195-4477-3/26/\$31.00~\copyright~2026 IEEE}
}

\title{Beyond Ground Truth: A Heuristic Unsupervised Quality Assurance Engine for Diagnosing Calibrational Blindness in Dental Segmentation}

\author{\IEEEauthorblockN{Fatematuj Johura Mim}

\IEEEauthorblockA{\textit{Department of Computer Science \& Engineering} \\
\textit{North East University Bangladesh}\\
 Sylhet, Bangladesh \\
  Email: fatematuj.jm@gmail.com}
}
\maketitle
\thispagestyle{firstpage}

\begin{abstract}
In real-world clinical settings, dental X-ray segmentation using deep learning methods suffers from reliability issues. Even when the segmentation quality is bad, deep learning models show high confidence in their predictions (calibrational blindness). Moreover, in real use, doctors do not have ground-truth masks to check their accuracy. To address this problem, we introduce a simple quality-checking engine called HUS-QA with three stages. We train it on the PRAD (Periapical Dental X-ray) dataset to build our SimCLR self-supervised ResNet-18 U-Net model, which generates the segmentation mask and then evaluates its quality. Our HUS-QA checks the segmentation mask quality without a ground-truth mask. Prediction confidence, edge sharpness, shape connectivity, and area-size penalty are the simple indicators considered in this research. Results show that standard confidence methods fail ($r = -0.1394$, $p = 1.7864 \times 10^{-3}$, $\text{ROC AUC} = 0.4275$). By contrast, our HUS-QA shows a positive correlation with actual Dice scores ($r = 0.4647, p = 3.7510 \times 10^{-28}$) and achieves a strong ROC-AUC of $0.8405$. Therefore, HUS-QA serves as a practical tool for clinical quality control.
\end{abstract}


\section{Introduction}
In modern medical systems, automatically segmenting dental X-rays is essential, as it helps clinicians diagnose oral conditions\cite{nooraldaim2023artificial, jader2018deep}. Deep learning models, especially encoder-decoder models \cite{ronneberger2015u}, help doctors detect oral diseases and save time by segmenting different dental anatomical structures in images \cite{litjens2017survey}. But in real-world use, \text{calibrational blindness} is a major problem. Deep learning models often show high confidence even when their output masks are wrong or incomplete \cite{guo2017calibration}. After deployment, clinicians have no labels to check whether the masks generated by deep learning models are reliable or misleading \cite{valindria2017reverse}.

We need a quality-control framework that can assess predicted mask quality without ground truth to build trust in Medical AI tools \cite{valindria2017reverse, azizi2021big}. Standard quality checks only look at basic confidence scores, but high confidence does not guarantee good shape or structure \cite{guo2017calibration}. To solve this issue, we present the \textbf{Heuristic Unsupervised Quality Assurance (HUS-QA)} engine. By looking at four simple visual indicators such as confidence score, boundary sharpness, component connectivity, and anatomical area size, HUS-QA  checks the segmentation quality of X-rays. It combines these indicators into one final score to tell doctors if a mask needs manual checking. Also, to lower the cost of expert annotations, our framework uses Self-Supervised Learning (SimCLR) to learn features from unannotated X-ray images first \cite{chen2020simple, chaitanya2020contrastive}.

The primary contributions of this work are:
\begin{itemize}
    \item We present HUS-QA, a simple approach that checks dental segmentation quality and catches overconfident errors without using ground-truth masks.
    \item We combine basic confidence scores with structural signals like Sobel edge sharpness, shape connectivity, and area penalties.
    \item We use a SimCLR self-supervised ResNet-18 \cite{he2016deep} U-Net model on the PRAD dataset to get good segmentation results with fewer labeled images.
    \item Test results show that HUS-QA has a positive correlation with Dice scores ($r = 0.4647$, $p = 3.7510 \times 10^{-28}$) and yields a strong ROC-AUC of $0.8405$ for separating bad segmentations, which is significantly better than standard confidence methods ($-0.1394$, $\text{ROC-AUC} = 0.4275$).
\end{itemize}

\section{Related Work}

\subsection{Deep Learning in Dental Radiography}
In medical image segmentation, convolutional neural networks such as U-Net models \cite{ronneberger2015u} are standard AI tools \cite{litjens2017survey}. Deep learning methods are widely used for tooth segmentation, cancer detection, caries detection, and anatomical structure analysis in the dental sector \cite{nooraldaim2023artificial, jader2018deep}. However, training traditional supervised models requires a huge amount of pixel-level annotated data from experts, which is time-consuming and expensive.

\subsection{Self-Supervised Learning in Medical AI}
The problem of labeled data scarcity can be solved by Self-Supervised Learning (SSL). Contrastive methods like SimCLR \cite{chen2020simple}  learn useful features from unlabeled images by pulling the two augmented views of a single image close together in feature space, while pushing views from other images further apart. The segmentation quality can be improved by SSL pre-trained models when labeled data is small in size \cite{chaitanya2020contrastive, azizi2021big}. Additionally, recent studies show that self-supervised learning has performed well in learning features without manual labeling in medical imaging\cite{ali2021self}.

\subsection{Quality Control and Calibrational Blindness}
Deep neural networks can show high confidence even if they do not predict correctly. They frequently give very high confidence on noisy data\cite{guo2017calibration}. In real medical systems, \textit{calibrational blindness} can be risky for human beings. To check the prediction quality of models without ground truth, Valindria et al. have proposed reverse classification accuracy (RCA) \cite{valindria2017reverse}. Calculating model uncertainty is used by other methods to predict segmentation quality \cite{devries2018leveraging} or fix calibration issues \cite{zou2025toward}. On the other hand, using uncertainty estimation methods can be slower or heavier in real-time use. Therefore, shape rules and boundary metrics can be used to identify bad outputs without heavy uncertainty estimation. Moving forward with the concept, our HUS-QA engine combines confidence, edge sharpness, connectivity, and area limits into a simple framework which is designed to address dental segmentation errors and assess the quality of the segmentation.

\textbf{Research Gap.} Some methods can check segmentation quality without ground truth, for example, reverse classification accuracy \cite{valindria2017reverse} and uncertainty-based methods \cite{devries2018leveraging, zou2025toward}. But these were mostly made for general medical images. As per our knowledge, it remains untested whether a model's confidence can be trusted as a quality score on dental X-rays, although overconfidence is a known problem \cite{guo2017calibration}. Our results show that this is a real problem. On periapical X-rays, confidence does not follow segmentation quality (AUC = 0.4275), and a nearly empty mask can still get 99.96\% confidence (Fig.~\ref{fig:qualitative_audit}, Case C). So we propose HUS-QA, a simple quality check that does not need ground-truth masks.

\section{Methodology}
\begin{figure}[htbp]
\centering
\includegraphics[width=1.05\linewidth]{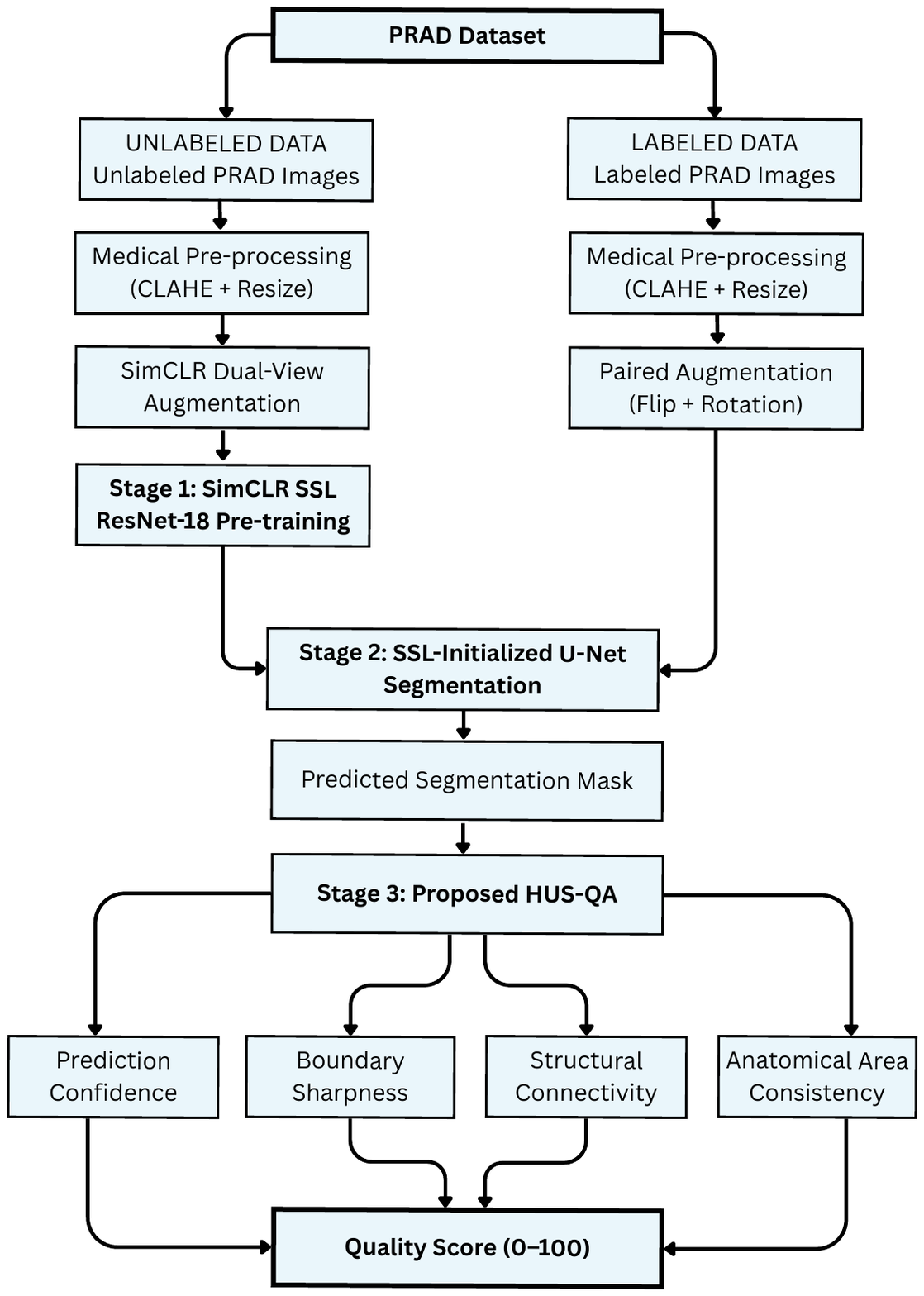}
\caption{Overall framework of the SimCLR ResNet-18 U-Net segmentation and HUS-QA quality check pipeline.}
\label{fig:arch}
\end{figure}

The full design of our three-stage framework is shown in Fig. \ref{fig:arch}. It works as an automatic quality assurance system using the PRAD dataset. During testing, a periapical dental X-ray is first pre-processed and sent through the SimCLR-initialized ResNet-18 \cite{he2016deep} U-Net to produce a tooth mask. Next, the HUS-QA module checks the output reliability without ground-truth masks and gives a quality score to help doctors accept or reject the result.

\subsection{Dataset Setup and Data Splitting:}
We evaluated our system (HUS-QA) on PRAD dataset with separate data splits for self-supervised pre-training and supervised training tasks. At first, we performed self-supervised feature learning using $2,500$ unlabeled X-ray images in Stage 1. For Stage 2, the labeled dataset was split into three sets. These three sets were $1,750$ image-mask pairs for training, $250$ pairs for validation, and $500$ pairs for final testing.

\subsection{Stage 1: Pre-processing and Self-Supervised Learning:}

\textbf{Image Pre-processing:} 
At first, we converted dental X-rays into grayscale and enhanced them with CLAHE where the clip limit is $2.0$, and the tile grid is $8 \times 8$ to make details clearer. Then images were saved as 3-channel RGB and resized to $224 \times 224$ pixels.

\textbf{SimCLR Augmentations:} 
Each image was transformed into two views using random crops ($\text{scale} = 0.6\text{--}1.0$), horizontal flips ($p = 0.5$), brightness adjustments ($p = 0.8$), ImageNet normalization, and Gaussian noise ($p = 0.3, \sigma = 0.05$).

\textbf{Self-Supervised Backbone Pre-training:} 
A ResNet-18 \cite{he2016deep} encoder with a two-layer projection head was trained on the $2,500$ unlabeled pre-training images using the SimCLR self-supervised learning method. The Normalized Temperature-scaled Cross-Entropy (NT-Xent) loss was used to optimize the encoder:

\begin{equation}
\mathcal{L}_{\text{NT-Xent}} = -\log \frac{\exp(\text{sim}(z_i, z_j)/\tau)}{\sum_{k=1}^{2N} \mathbb{I}_{[k \neq i]} \exp(\text{sim}(z_i, z_k)/\tau)}
\end{equation}

Here, $z_i$ and $z_j$ are the projected representations of the two augmented views. $\text{sim}(\cdot)$ denotes cosine similarity, and $\mathbb{1}_{[k \neq i]}$ is an indicator function that equals 1 when $k \neq i$. $\tau$ is the temperature parameter, set to 0.5 in this work. The encoder was pre-trained for 100 epochs to learn robust structural representations from the unlabeled periapical dental radiographs.

\subsection{Stage 2: Dental segmentation initiated by Self-Supervised Learning:}
 The pre-trained SimCLR ResNet-18 encoder was used as the encoder of a U-Net architecture for supervised dental segmentation. We applied standard medical preprocessing, including CLAHE, to all input images to enhance boundary clarity and resized them to $224 \times 224$ pixels. The SimCLR self-supervised ResNet-18 U-Net model was trained on $1,750$ paired images and masks, with model selection guided by the $250$ validation pairs. For 25 training epochs, the model received paired data augmentation in the form of random horizontal flipping and rotation ($\pm 15^\circ$).

The segmentation network was optimized using a hybrid Dice-BCE loss:

\begin{equation}
\mathcal{L} = \lambda \mathcal{L}_{\text{BCE}} + (1 - \lambda) \mathcal{L}_{\text{Dice}}
\end{equation}

where $\lambda = 0.5$. The AdamW optimizer with cosine annealing learning-rate scheduling was used to train the network. The model, which achieved the highest validation Dice score on the validation set, was selected for the subsequent quality assessment stage.

\subsection{Stage 3: Proposed Ground-Truth-Free Segmentation Quality Assessment (HUS-QA):}
Without relying on ground-truth annotations, the proposed HUS-QA module assesses the predicted segmentations across $500$ test samples. To estimate overall reliability, it analyzes both the soft prediction confidences and the corresponding binary masks.

Four complementary quality indicators are extracted:
\begin{itemize}
    \item \textbf{Prediction Confidence ($Q_c$):} Computed from the confidence distribution of the predicted soft mask.
    \item \textbf{Boundary Sharpness ($Q_s$):} Estimated using the Sobel gradient magnitude of the predicted segmentation boundaries.
    \item \textbf{Structural Connectivity ($Q_{\text{str}}$):} Measured from the connected components of the binary segmentation mask to penalize fragmented or isolated predictions.
    \item \textbf{Anatomical Area Consistency ($Q_{\text{area}}$):} An area penalty factor to suppress anatomically unrealistic segmentation regions.
\end{itemize}

The first three indicators are combined to compute the geometric quality score:

\begin{equation}
Q_g = 0.40 Q_c + 0.40 Q_s + 0.20 Q_{\text{str}}
\end{equation}

We chose the weights ($0.40, 0.40, 0.20$) based on $250$ validation pairs only. Edge sharpness ($Q_s$) and confidence ($Q_c$) were the main factors of the mask quality, which are equally important ($0.40$). Shape connectivity ($Q_{\text{str}}$) provided a supporting check, with a lower weight ($0.20$).

Finally, our HUS-QA score was obtained by including the anatomical area penalty:

\begin{equation}
Q_{\text{final}} = Q_g \times P_{\text{area}}
\end{equation}

 The range was limited to  $0\text{--}100$ for the resulting score, where higher values indicate more reliable segmentation outputs. During testing, HUS-QA automatically predicted segmentation quality without requiring manual annotations, providing an effective quality assessment mechanism for clinical support.

\section{Results and Discussion}

\subsection{Segmentation Performance}
We used $500$ test images from the PRAD dataset to evaluate the SimCLR ResNet-18 U-Net model. As reported in Table \ref{tab:seg_performance}, the model acquired an Accuracy of $0.9872$, a Dice Score of $0.6961$, and an IoU (Jaccard) of $0.5666$.

We intentionally did not try to improve the segmentation model to perfection. Instead, we wanted to test our HUS-QA engine on challenging and imperfect masks. This proves that our QA engine can successfully detect and measure errors even when the segmentation model makes mistakes.

\begin{table}[H]
\centering
\caption{Segmentation Performance on the PRAD Test Dataset}
\label{tab:seg_performance}
\begin{tabular}{|l|c|}
\hline
\textbf{Metric} & \textbf{Score} \\ \hline
Accuracy & $0.9872$ \\ \hline
Dice Score & $0.6961$ \\ \hline
IoU (Jaccard) & $0.5666$ \\ \hline
\end{tabular}
\end{table}

\subsection{Segmentation Quality Assessment}

\subsubsection{Baseline Quality Assessment}
\begin{figure}[H]
\centering
\includegraphics[width=\linewidth]{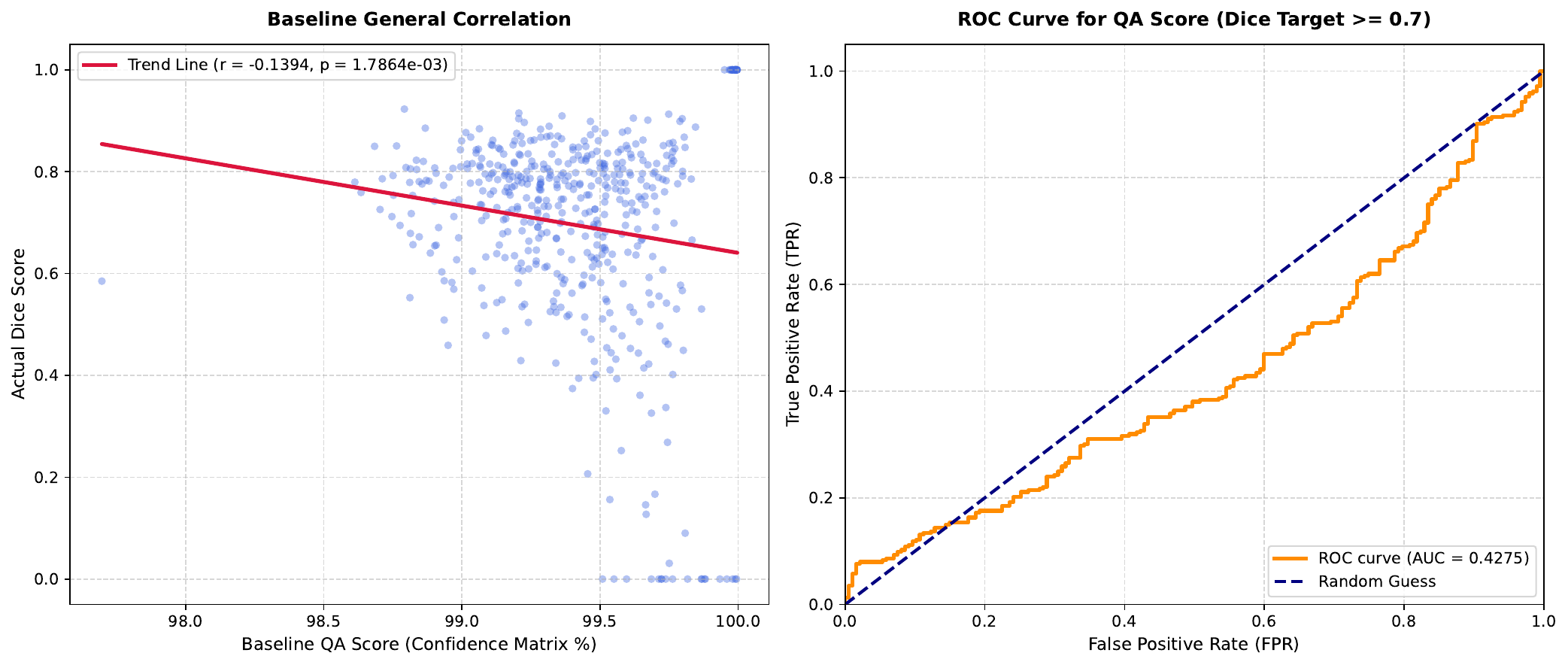}
\caption{Quality checking performance using standard confidence baseline.}
\label{fig:baseline_qa}
\end{figure}

Fig. \ref{fig:baseline_qa} illustrates the performance of the baseline confidence-based quality assessment method on the test set. We considered a segmentation acceptable when its Dice score reached $70\%$ or higher, following the literature \cite{taha2015metrics,zou2004statistical}. This cutoff was used to label samples as acceptable or poor for the ROC-AUC analysis in Fig.~\ref{fig:baseline_qa} and Fig.~\ref{fig:hus_qa}. The scatter plot reveals a weak negative correlation ($r = -0.1394, p = 1.7864 \times 10^{-3}$) with the ground-truth Dice score. The corresponding ROC curve yields an AUC of only $0.4275$, performing worse than random guessing ($\text{AUC} = 0.50$). This happens because of how the model behaves when it predicts nothing. As Fig.~\ref{fig:qualitative_audit} (Case C) shows, when the model predicts an almost empty mask, its confidence stays very high, since the sigmoid values are close to zero across the whole image. So even the worst predictions can get high confidence scores. This is why the correlation turns negative, and the AUC ends up below random guessing.

\subsubsection{Proposed HUS-QA Performance}
\begin{figure}[H]
\centering
\includegraphics[width=\linewidth]{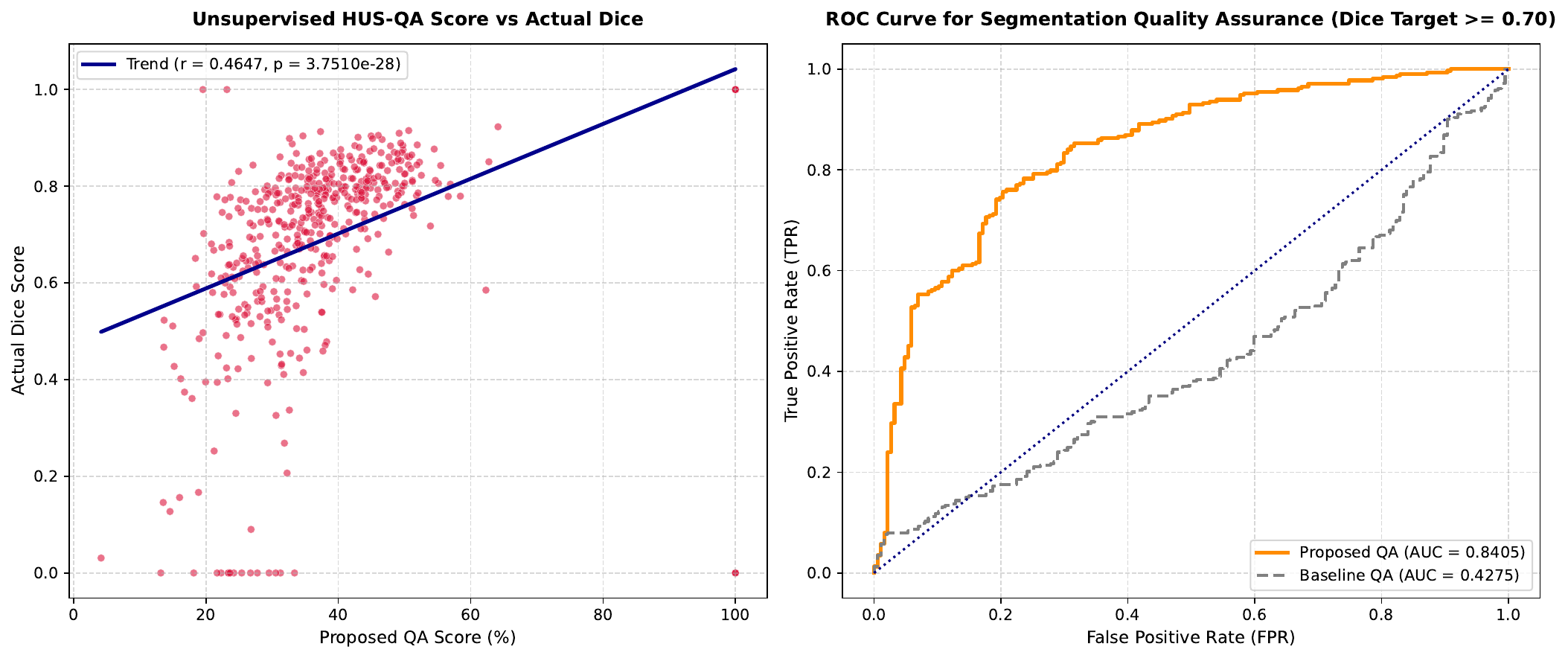}
\caption{Quality checking performance using the proposed HUS-QA framework.}
\label{fig:hus_qa}
\end{figure}

Fig. \ref{fig:hus_qa} highlights the evaluation of the proposed HUS-QA engine. HUS-QA showed a positive correlation with actual Dice scores ($r = 0.4647$), with a very small $p$-value ($p = 3.7510 \times 10^{-28}$). The ROC curve gives a strong AUC of $0.8405$, which is much better than the baseline AUC of $0.4275$. Integrating edge sharpness, connectivity, and anatomical area constraints allows HUS-QA to reliably diagnose faulty outputs without ground truth.

\begin{table}[H]
\centering
\caption{Quality Assessment Performance Comparison}
\label{tab:qa_performance}
\begin{tabular}{|l|c|c|c|}
\hline
\textbf{Method} & \textbf{Pearson ($r$)} & \textbf{$p$-value} & \textbf{ROC AUC} \\ \hline
Baseline QA & $-0.1394$ & $1.7864 \times 10^{-3}$ & $0.4275$ \\ \hline
Proposed HUS-QA & $\mathbf{0.4647}$ & $\mathbf{3.7510 \times 10^{-28}}$ & $\mathbf{0.8405}$ \\ \hline
\end{tabular}
\end{table}

During testing and evaluation, the proposed HUS-QA framework assessed the segmentation quality of 500 untouched test samples without using their ground-truth masks.

\subsection{Threshold Sensitivity Analysis}
In our main evaluation, we considered a segmentation acceptable when its Dice score reached $70\%$ or higher. To check that this specific cutoff does not bias the reported results, we recomputed the ROC-AUC for both the baseline QA method and the proposed HUS-QA across a range of thresholds (0.60 to 0.80). Table \ref{tab:sensitivity} shows the results.

\begin{table}[H]
\centering
\caption{ROC-AUC Across Different Dice Thresholds}
\label{tab:sensitivity}
\begin{tabular}{|c|c|c|c|c|}
\hline
\textbf{Dice} & \textbf{HUS-QA} & \textbf{Baseline} & \textbf{N} & \textbf{N} \\
\textbf{Threshold} & \textbf{AUC} & \textbf{QA AUC} & \textbf{(Acc.)} & \textbf{(Poor)} \\ \hline
0.60 & 0.8279 & 0.3461 & 390 & 110 \\ \hline
0.65 & 0.8397 & 0.3957 & 355 & 145 \\ \hline
0.70 & 0.8405 & 0.4275 & 313 & 187 \\ \hline
0.75 & 0.8238 & 0.4796 & 259 & 241 \\ \hline
0.80 & 0.8264 & 0.5506 & 157 & 343 \\ \hline
\end{tabular}
\end{table}

The AUC of HUS-QA stays close to 0.82--0.84 across all thresholds tested, even though the class balance changes considerably (from 390:110 at 0.60 to 157:343 at 0.80). This suggests the reported performance is not tied to the specific 0.7 cutoff. The baseline QA method stays weak at every threshold, with AUC ranging from 0.3461 to 0.5506 -- close to random guessing (AUC = 0.50) and far below HUS-QA at all points tested.

\begin{figure}[H]
\centering
\includegraphics[width=\linewidth]{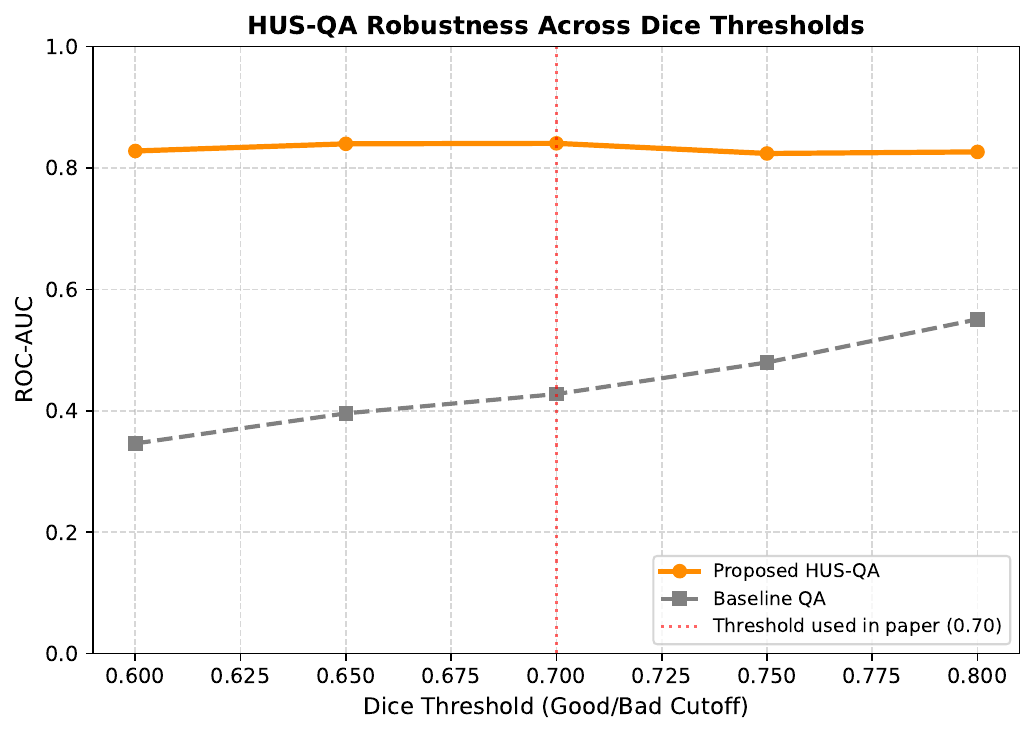}
\caption{ROC-AUC of HUS-QA and baseline QA across Dice thresholds from 0.60 to 0.80. The red dotted line marks the 0.70 threshold used in the main evaluation.}
\label{fig:sensitivity}
\end{figure}

\subsection{Qualitative Analysis}
\begin{figure}[H]
\centering
\includegraphics[width=\linewidth]{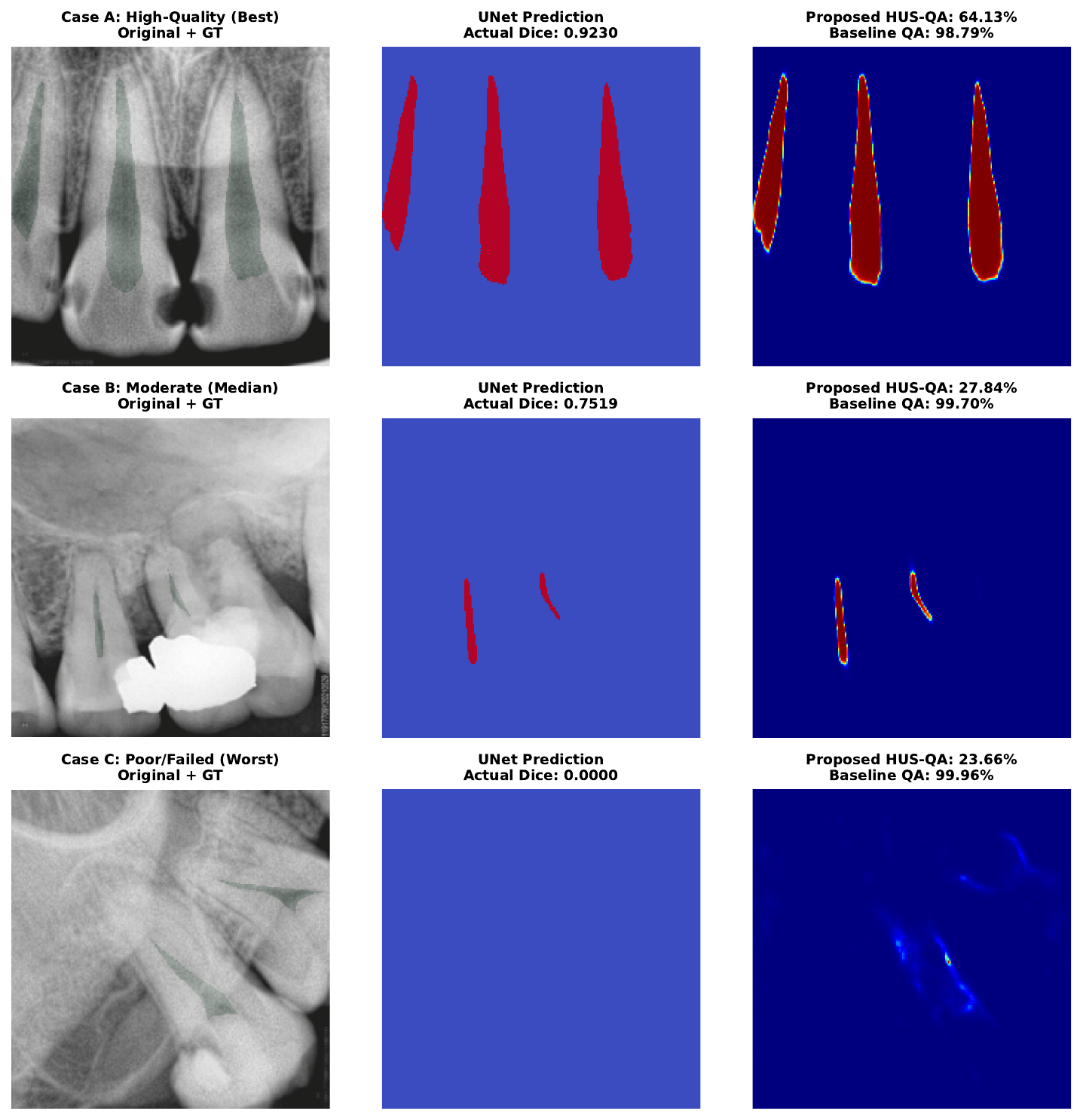}
\caption{Visual comparison of segmentation outputs under three quality levels.}
\label{fig:qualitative_audit}
\end{figure}

Fig. \ref{fig:qualitative_audit} provides a visual comparison across three qualitative regimes. For high-quality segmentations (Case A), the predicted mask aligns closely with anatomical boundaries, giving high Dice and HUS-QA scores. In moderate failure cases (Case B), HUS-QA scores drop considerably due to severe under-segmentation and anatomical area loss. In severe failure cases (Case C), HUS-QA produces very low scores, marking these predictions for clinician audit. Conversely, the baseline confidence score remains uncalibrated, yielding high confidence values across all failure modes.

\FloatBarrier

\subsection{Discussion}
Our findings confirm that incorporating structural, boundary, and geometrical priors addresses calibrational blindness in deep learning segmentations without reference masks. HUS-QA exhibits superior correlation ($r = 0.4647$) compared to naive confidence ($r = -0.1394$), providing a reliable safeguard for automated dental radiography diagnostics. Although we tested our method only on the PRAD dataset, HUS-QA uses basic shape and edge rules. These general rules do not depend on a specific dataset, so the method may also work for other types of medical images.

\section{Conclusion}
This study proposed HUS-QA, an unsupervised quality assurance framework designed to detect overconfident segmentation errors in dental AI workflows. Evaluated on dental X-rays, HUS-QA synthesizes confidence, boundary sharpness, structural connectivity, and anatomical constraints without ground-truth masks. The framework achieves a moderate Pearson correlation with actual Dice scores ($r = 0.4647, p = 3.7510 \times 10^{-28}$) and an ROC AUC of $0.8405$, outperforming standard confidence baselines ($r = -0.1394, \text{ROC AUC} = 0.4275$).

As mentioned in Section IV-A, we chose a moderately performing model on purpose to test HUS-QA under difficult conditions. In future work, we plan to test HUS-QA on a better-performing U-Net to see if it still works well when segmentation quality is high. We also plan to extend the indicator set and test HUS-QA on panoramic radiographs and 3D CBCT imaging techniques.

\end{document}